\documentclass[11pt]{article}

\usepackage[preprint]{acl}

\usepackage{times}
\usepackage{latexsym}
\usepackage{enumitem}
\usepackage[T1]{fontenc}

\usepackage[utf8]{inputenc}

\usepackage{microtype}

\usepackage{inconsolata}

\usepackage{graphicx}
\graphicspath{{fig/}}

\usepackage{booktabs}                  
\usepackage{multirow}                  
\usepackage{subcaption}                
\usepackage{tikz}                      
\usetikzlibrary{arrows.meta,backgrounds,fit,positioning}
\usepackage{amsmath,amssymb,amsthm}    
\usepackage{bm}                        
\usepackage{algorithm}                 
\usepackage{algpseudocode}             
\usepackage{xcolor}                    
\usepackage[most]{tcolorbox}           
\usepackage{fvextra}                   
\fvset{breaksymbolleft={},breaksymbolright={}} 
\usepackage{cleveref}                  

\newtcolorbox{promptbox}[1]{%
  breakable, enhanced,
  colback=gray!4, colframe=black!50,
  arc=2pt, boxsep=4pt,
  left=6pt, right=6pt, top=4pt, bottom=4pt,
  title=#1, fonttitle=\bfseries\small,
}

\newcommand{\skillinjector}{SkillsInjector}

\title{\skillinjector: Dynamic Skill Context Construction for LLM Agents}

\author{
  \textbf{Yanchao Li}\textsuperscript{1,2},
  \textbf{Wanhao Liu}\textsuperscript{2},
  \textbf{Ben Gao}\textsuperscript{2},
  \textbf{Jiaqing Xie}\textsuperscript{2}, \\
  \textbf{Zhehong Ai}\textsuperscript{2\,$\dagger$},
  \textbf{Na Zou}\textsuperscript{2},
  \textbf{Yuqiang Li}\textsuperscript{2},
  \textbf{Tianfan Fu}\textsuperscript{1,2\,$\dagger$}
  \\
  \textsuperscript{1}Nanjing University \quad
  \textsuperscript{2}Shanghai AI Lab
}

\begin{document}
\maketitle
\begin{abstract}
LLM agents now draw on growing skill libraries to handle complex tasks.
However, injecting more skills does not always improve task completion and can even degrade it.
Existing methods still treat skill injection as a static step, selecting skills with fixed criteria, fixing the budget in advance, and leaving descriptions unchanged.
We argue that this static treatment can undermine the utility of skills, because which skills are exposed, how many are included, and how they are presented all affect downstream performance.
We propose \skillinjector{}, a two-stage adaptive method that jointly addresses these decisions.
First, a context planner learns execution-grounded skill preferences and admits an adaptive number of skills for each task.
A set-aware renderer then tailors how selected descriptions are presented relative to their co-injected neighbors.
Across tau2-bench, SkillsBench, and ALFWorld, \skillinjector{} achieves the highest score, improving over the strongest baseline by 3.9, 6.1, and 7.3 percentage points, respectively.
Ablation studies show that skill selection, adaptive budgeting, and set-aware rendering each contribute to the gain.
These results show that skill-augmented agents benefit from optimizing the injected context itself.
Code will be released upon publication
\end{abstract}

\section{Introduction}
\label{sec:intro}

\begin{figure*}[t]
\centering
\includegraphics[width=\textwidth]{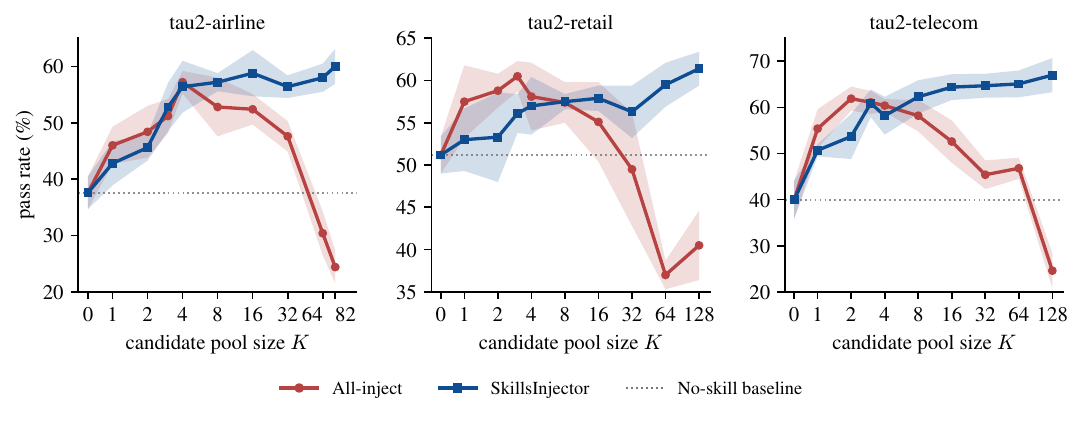}
\caption{\textbf{Scaling the candidate skill pool on tau2-bench.}
Static all-injection (red) collapses as the candidate skill pool grows, while our method (blue) remains stable. }
\label{fig:scaling}
\end{figure*}

Large Language Model (LLM) agents increasingly rely on skill injection to extend their capabilities across complex and open-ended tasks~\citep{li2026skillsbenchbenchmarkingagentskills, zhou2026mementoskillsletagentsdesign, jiang2026sokagenticskills}. Intuitively, exposing a larger library of specialized skills should empower an agent to handle more sophisticated scenarios. However, recent empirical evidence points to a scaling bottleneck~\citep{liu-etal-2024-lost, gao2026skillreduceroptimizingllmagent, liu2026agenticskillsworkwild}. Packing more skills can degrade downstream performance rather than improving it (Figure~\ref{fig:scaling}). As the injected skill set grows, agents can suffer from attention dispersion and competition among similar skills~\citep{li2026skillsbenchbenchmarkingagentskills, li2026singleagentskillsreplacemultiagent}. While the intrinsic quality of underlying skills remains fundamental, the suboptimal process of skill injection itself emerges as a primary bottleneck restricting advanced agent systems.

Existing methods address this bottleneck by optimizing isolated pipeline stages.
Retrieval and routing methods rank candidate skills and expose a fixed-size shortlist~\citep{zheng2026skillrouterskillroutingllm, molfetta-etal-2025-ports}. 
A separate line evolves or curates the skill library itself~\citep{zhou2026mementoskillsletagentsdesign, xia2026skillrlevolvingagentsrecursive}.
These directions are useful, but they still treat skill context generation as a static step. 
The budget is fixed in advance, and each skill description is passed to the agent unchanged.
We argue that the missing layer is task-adaptive skill context construction, which jointly decides which skills to expose, how many to include, and how to present them together for the current task.

To bridge this gap, we view skill injection as dynamic, task-adaptive construction of the skill context rather than static retrieval. 
We propose \skillinjector{}, a framework that jointly addresses skill selection, budgeting, and presentation through two coordinated components: a context planner and a set-aware renderer.
The context planner estimates the execution-grounded utility of each candidate skill and admits an adaptive set of high-benefit skills for the current task, rather than relying only on semantic similarity or a fixed rule.
The selected set is then passed to the renderer, which adapts each skill description in relation to its co-injected neighbors, clarifying role boundaries and reducing overlap.
This design provides the agent with a more focused skill context at low inference cost.

Across tau2-bench, SkillsBench, and ALFWorld, \skillinjector{} achieves the strongest results, and ablations show that selection, budgeting, and rendering each contribute.
Further diagnostics characterize the structure behind the skill-injection bottleneck.
Skill effects are heterogeneous across both quality and cost, and tasks differ in how much skill context they can exploit.
These findings motivates task-specific context allocation that selects compatible skills and makes their roles explicit.

Our contributions are as follows.
\begin{itemize}[leftmargin=*]
  \item We reframe skill injection as a per-task context construction problem, shifting the focus from skill matching alone to constructing the context the agent should read.
  \item We introduce \skillinjector{}, which couples an execution-grounded planner for adaptive selection and budgeting with a set-aware renderer for description contextualization.
  \item We evaluate \skillinjector{} across tau2-bench, SkillsBench, and ALFWorld, where it consistently surpasses classical retrieval and recent skill-routing methods. Our diagnostics further show why skill selection, adaptive budgeting, and description presentation are all necessary.
\end{itemize}
\section{Related Work}
\label{sec:related}

\paragraph{Skill libraries for LLM agents.}
Skill libraries are now a standard layer in LLM agent systems~\citep{jiang2026sokagenticskills, li2026skillsbenchbenchmarkingagentskills}.
One line of work evolves the library itself, distilling new skills from execution traces or co-evolving them with the agent policy~\citep{zhou2026mementoskillsletagentsdesign, xia2026skillrlevolvingagentsrecursive, yang2026autoskillexperiencedrivenlifelonglearning}.
A closely related line keeps the executor frozen and trains a separate module to curate which skills are applied~\citep{ouyang2026skilloslearningskillcuration}.
These directions all change the library itself.
We instead take the library as fixed, as in most real deployments.
Our focus is how to assemble its skill context for each task.

\paragraph{Retrieval over tools and skills.}
Most existing work treats skill injection as a retrieval problem.
Classical tool-augmented agents rank skills by documentation similarity~\citep{NEURIPS2024_e4c61f57, ICLR2024_28e50ee5}, and skill-oriented retrievers extend this with listwise reranking or dependency-aware graph structure~\citep{zheng2026skillrouterskillroutingllm, liu2026graphskillsdependencyawarestructural}.
A separate thread lets the model emit tool identifiers directly~\citep{NEURIPS2023_8fd1a81c, ICLR2025_b646bdeb}.
All these methods fix the budget in advance and pass each description unchanged.
We let the budget vary with the task, and the renderer refines each description against its co-injected neighbors.

\paragraph{Improving skill descriptions.}
What an agent sees of each skill shapes which one it picks.
Empirical audits show catalog descriptions are systematically defective and lose much of their benefit under realistic noise~\citep{hasan2026modelcontextprotocolmcp, liu2026agenticskillsworkwild}.
One line of work rewrites descriptions offline through task-agnostic prompting or trial-and-error refinement~\citep{yuan-etal-2025-easytool, fang-etal-2025-play2prompt}.
A complementary line updates descriptions after agent failures~\citep{zhou2026mementoskillsletagentsdesign, ghoshal2026jtprojointtoolpromptreflective}.
All these methods produce a static description per skill.
Our renderer instead adapts at injection time.
It conditions each description on the specific set of co-injected skills.

\section{Problem Formulation}

\begin{figure*}[t]
  \centering
    \includegraphics[width=\textwidth]{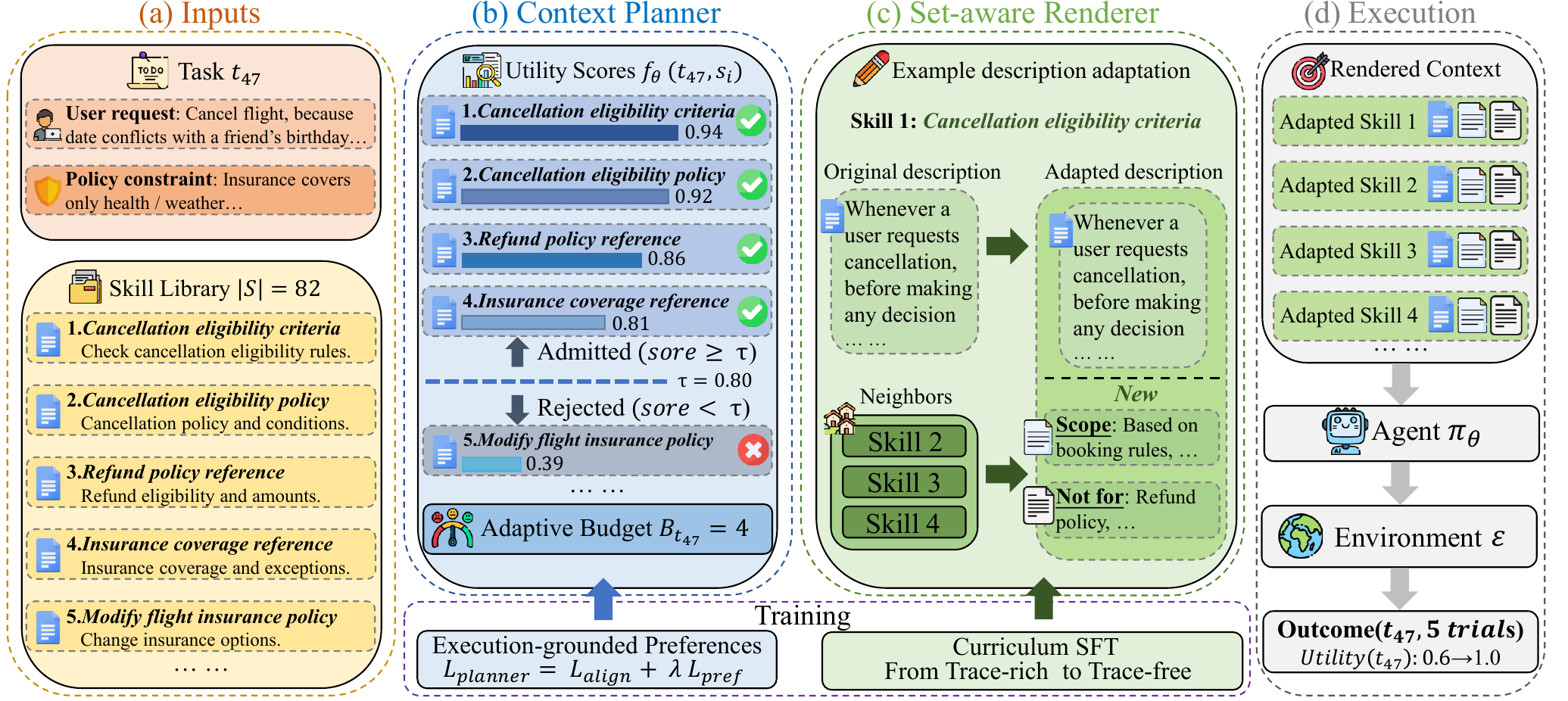}
  \caption{\textbf{\skillinjector{} pipeline.}
  The planner learns execution-grounded skill utility, while the renderer adapts selected descriptions with awareness of co-injected skills.
  At inference, the trained components select an adaptive skill set and inject the rendered context into a frozen agent.}
  \label{fig:framework}
\end{figure*}

\label{sec:problem}

We consider skill injection for a frozen agent operating over a fixed skill library.
A task $t \sim \mathcal{D}$ is executed in environment $\mathcal{E}$ by policy $\pi_\theta$, producing a trajectory $\tau$ with success indicator $r(\tau) \in \{0, 1\}$.
The library is $\mathcal{S} = \{s_1, \ldots, s_N\}$, and each skill $s_i = (d_i, b_i)$ pairs an agent-visible description $d_i$ with full procedural content $b_i$.
At inference the policy and library are fixed, and the controlled object is the skill context supplied to the agent.
To make these objects concrete, we use one tau2-airline task $t_{47}$ as a running example throughout the paper.
In $t_{47}$, the user asks the agent to cancel a flight because the date conflicts with a friend's birthday, while the airline's insurance covers only health or weather reasons.
The relevant library has $|\mathcal{S}| = 82$ airline policy-reference skills.

For a task $t$ and an injected skill context $\tilde{\mathcal{C}}$, we measure the value of $\tilde{\mathcal{C}}$ by the expected task completion of the frozen agent,
\begin{equation}
\label{eq:utility}
\mathcal{U}(t, \tilde{\mathcal{C}}) = \mathbb{E}_{\tau \sim \pi_\theta(\cdot \mid t, \tilde{\mathcal{C}}, \mathcal{E})}[r(\tau)].
\end{equation}
This utility evaluates a context by execution outcome rather than by surface signals such as semantic similarity.
The execution-grounded benefit of skill $s$ for task $t$ is $\Delta(t, s) = \mathcal{U}(t, \{s\}) - \mathcal{U}(t, \emptyset)$, the gain over a no-skill baseline.
For $t_{47}$, $\mathcal{U}(t_{47}, \emptyset) = 0.60$ over five seeds, and $\Delta(t_{47}, s)$ ranges from $-0.20$ to $+0.40$ across the $82$ candidate skills.

Skill context construction is a mapping $A(t, \mathcal{S}) = (\mathcal{C}_t, \rho_t)$, the context-construction operator, producing a selected subset $\mathcal{C}_t \subseteq \mathcal{S}$ and a rendering function $\rho_t$ that adapts each description while leaving each body $b_i$ unchanged.
The induced budget is $B_t = |\mathcal{C}_t|$ and the rendered context is $\tilde{\mathcal{C}}_t = \rho_t(\mathcal{C}_t)$.
We seek $A$ that maximizes expected execution utility $\mathbb{E}_{t \sim \mathcal{D}}[\mathcal{U}(t, \tilde{\mathcal{C}}_t)]$ over the task distribution.
For $t_{47}$, $A$ returns $B_{t_{47}} = 4$ skills with rendered descriptions. Standard retrieval is the special case of this mapping.
Our formulation generalizes retrieval in that (i) $B_t$ varies with $t$, (ii) $\rho_t$ depends on the selected set $\mathcal{C}_t$, and (iii) selection targets $\Delta(t, s)$ directly.

\section{Method}
\label{sec:method}
\label{sec:method:procedure}

We instantiate the context-construction operator $A(t, \mathcal{S})$ as \skillinjector{}.
The first component is a context planner.
It handles selection and budget, because tasks differ in how many useful skills they need and skills with similar descriptions can behave differently.
The second component is a set-aware renderer.
It handles the presentation decision, since descriptions can overlap or shadow each other under co-injection.
Both components are trained on a training task pool.
Figure~\ref{fig:framework} summarizes the design, and Algorithm~\ref{alg:pipeline} in Appendix~\ref{app:method-details} gives the full training and inference pipeline.

\subsection{Context Planner}
\label{sec:method:planner}

The planner ranks candidate skills for a task and decides how many enter the context.
Semantic similarity alone is not enough, because surface relevance does not predict whether a skill helps the agent.
For example, for $t_{47}$, the candidate \texttt{modify\_flight\_insurance\_policy} shares the insurance keyword with the user's request yet has $\Delta = -0.20$, while four policy-reference skills yield $\Delta = +0.40$.
We therefore treat planning as a preference learning problem, using the execution-grounded benefit $\Delta(t, s)$ from Section~\ref{sec:problem} as the supervision signal.
A scorer $f_\theta(t, s)$ is trained to predict $\Delta(t, s)$ from the task and description.
For generality we let $f_\theta$ act on a per-task candidate set $\mathcal{S}_c(t) \subseteq \mathcal{S}$.
See Appendix~\ref{app:method-details} for more details.

\paragraph{Scoring.}
The supervision from $\Delta(t, s)$ has two useful structures.
It forms a soft distribution over relative benefit, and within that distribution it provides pairwise orderings between skills.
We bound $\Delta(t, s)$ to a label $y(t, s) \in [0, 1]$ and turn it into a benefit distribution over $\mathcal{S}_c(t)$ with temperature $\beta$,
\begin{equation}
\label{eq:q-delta}
q_\Delta(s \mid t) = \frac{\exp(y(t, s) / \beta)}{\sum_{s' \in \mathcal{S}_c(t)} \exp(y(t, s') / \beta)}.
\end{equation}
The scorer induces a predicted distribution $p_\theta(s \mid t)$ by an analogous softmax with temperature $\gamma$, and the first training term aligns it with $q_\Delta$,
\begin{equation}
\label{eq:align}
\mathcal{L}_{\mathrm{align}}(t) = \mathrm{KL}\!\left( q_\Delta(\cdot \mid t) \,\|\, p_\theta(\cdot \mid t) \right).
\end{equation}
Distribution alignment alone can blur fine-grained orderings among close skills, so we add a pairwise preference term over pairs ordered by $y$~\citep{10.1145/1102351.1102363}.
We use an odds-ratio form~\citep{hong-etal-2024-orpo}, with $o_\theta(s \mid t) = \log\!\left( p_\theta(s \mid t) / (1 - p_\theta(s \mid t)) \right)$, and define
\begin{equation}
\label{eq:pref}
\mathcal{L}_{\mathrm{pref}}(t) = \mathbb{E}_{(s^+, s^-)}\bigl[ \log\bigl( 1 + e^{o_\theta(s^- \mid t) - o_\theta(s^+ \mid t)} \bigr) \bigr],
\end{equation}
where $(s^+, s^-)$ ranges over pairs with $y(t, s^+) > y(t, s^-)$.
Negatives are drawn from the most semantically similar candidates in $\mathcal{S}_c(t)$ that the labels do not prefer, so the scorer learns to separate functionally close skills rather than topically unrelated ones.
The full planner objective is $\mathcal{L}_{\mathrm{plan}} = \mathbb{E}_t [\mathcal{L}_{\mathrm{align}}(t) + \lambda \mathcal{L}_{\mathrm{pref}}(t)]$, with $\lambda$ chosen on a held-out dev set.
After training, $f_\theta(t, s)$ supplies a per-task score for every candidate skill.

\paragraph{Budget.}
\label{sec:method:budget}
Scores from $f_\theta$ must be converted into a set rather than a ranking, but a fixed top-$K$ rule forces every task into the same size.
We first normalize scores within each task to the unit interval, which makes a single threshold comparable across tasks.
The selected set then contains every candidate whose normalized score exceeds a per-domain threshold,
\begin{equation}
\label{eq:budget}
\mathcal{C}^*_t = \left\{ s \in \mathcal{S}_c(t) \,:\, \tilde{f}_\theta(t, s) \geq \tau^\star_d \right\}, \quad B_t = \lvert \mathcal{C}^*_t \rvert.
\end{equation}
The threshold $\tau^\star_d$ is chosen per domain on a held-out dev set to maximize execution utility.
When no skill clears the threshold, we inject none.
The budget $B_t$ thus emerges from the task rather than being fixed in advance.
For example, for $t_{47}$, $f_\theta$ scores the four useful skills between $0.81$ and $0.94$ and the insurance distractor at $0.39$, and the threshold $\tau^\star_d$ admits the four while rejecting the distractor.

\subsection{Set-Aware Renderer}
\label{sec:method:renderer}

The renderer handles the presentation decision after the planner produces a selected set $\mathcal{C}^*_t$.
We distill set-aware adaptation into a small renderer $R_\phi$ called once before each agent invocation, avoiding online inference of a large model~\citep{hinton2015distillingknowledgeneuralnetwork}.
When $\lvert \mathcal{C}^*_t \rvert \leq 1$ the renderer is bypassed, since a single injected skill has no neighbors to disambiguate against.

\paragraph{Conditioning.}
For $\lvert \mathcal{C}^*_t \rvert \geq 2$, the renderer reads each selected description $d_i$, the task $t$, and the neighbor descriptions $\{d_j : j \neq i\}$, and produces an adapted $\tilde{d}_i$.
Only $d_i$ is adapted while the skill body $b_i$ passes through unchanged.
For $t_{47}$, $R_\phi$ adapts \texttt{cancellation\_eligibility\_criteria} by appending a scope clause referencing the co-selected neighbors, such as \texttt{Not for: refund\_policy\_reference, compensation\_eligibility\_criteria}.

\paragraph{Curriculum.}
Training $R_\phi$ requires set-aware adaptation supervision that the small renderer cannot produce on its own.
We construct it with a stronger teacher Qwen3-235B-A22B-Instruct-2507.
Starting from each original description $D_0 = d_i$, the teacher first produces a task-agnostic cleanup $D_1$ that sharpens wording without considering co-injection, then a set-aware target $D_2$ conditioned on the task, the selected neighbors, and a trace summary from a training rollout of $s_i$, which serves as the SFT target for the student.
Because trace summaries are unavailable at inference, we mix trace-rich and trace-free inputs via a curriculum~\citep{10.1145/1553374.1553380}.
The student is trained to predict $D_2$ from the mixed input,
\begin{equation}
\label{eq:rend}
\mathcal{L}_{\mathrm{rend}}(\phi) = \mathbb{E}_{(t, s_i, \mathcal{C}^*_t)}\bigl[ -\log p_\phi( D_2 \mid x_{t, i} ) \bigr],
\end{equation}
where $x_{t, i}$ is the trace-rich or trace-free input.
At inference $R_\phi$ reads only the task and the selected descriptions, and $\tilde{\mathcal{C}}^*_t$ is injected once into the frozen agent.

\begin{table*}[t]
\centering
\small
\setlength{\tabcolsep}{6pt}
\begin{tabular}{lcccccc}
\toprule
Method & tau2-airline & tau2-retail & tau2-telecom & SkillsBench & ALFWorld & Avg \\
\midrule
No-skill                                  & 37.6 & 51.2 & 40.0 & 5.2  & 67.1 & 40.2 \\
Random-skill                              & 42.4 & 53.0 & 41.9 & 6.7  & 69.0 & 42.6 \\
Full-library                              & 24.4 & 40.5 & 24.6 & 3.2  & 31.5 & 24.8 \\
BM25                                      & 43.6 & 54.9 & 51.2 & 12.8 & 71.2 & 46.7 \\
Dense Cosine                              & 45.2 & 55.3 & 54.7 & 14.2 & 72.9 & 48.5 \\
LLM-as-selector                           & 49.6 & 55.1 & 55.8 & 14.2 & 73.8 & 49.7 \\
SkillRouter~\citep{zheng2026skillrouterskillroutingllm}  & 54.0 & 59.8 & \underline{62.8} & \underline{16.5} & 74.4 & 53.5 \\
Graph of Skills~\citep{liu2026graphskillsdependencyawarestructural}       & \underline{56.1} & \underline{60.0} & 60.4 & 15.9 & \underline{75.4} & \underline{53.6} \\
\midrule
\skillinjector{} (Ours)                   & \textbf{60.0} & \textbf{61.4} & \textbf{67.0} & \textbf{22.6} & \textbf{82.7} & \textbf{58.7} \\
\bottomrule
\end{tabular}
\caption{\textbf{Main results.}
Task pass rate (\%, $\uparrow$) is reported across tau2-bench, SkillsBench, and ALFWorld.
\skillinjector{} achieves the best score on every benchmark. The \textbf{best} and \underline{second-best} results are highlighted, respectively. }
\label{tab:main}
\end{table*}

\section{Experiment}

\subsection{Experimental Setup}
\label{sec:experiments}

To cover different skill regimes, we evaluate \skillinjector{} on tau2-bench (three customer-service domains), SkillsBench (packaged various skills), and ALFWorld (interactive decision-making).
Within each benchmark, the agent policy $\pi_\theta$ and the harness are held fixed across compared methods, and the only controlled variable is the injected skill context.

\paragraph{Benchmarks.}
\textit{tau2-bench}~\citep{barres2025tau2benchevaluatingconversationalagents} contains three customer-service domains (airline, retail, telecom) with 50, 114, and 114 tasks respectively. 
Because the three domains differ substantially, we report results per-domain throughout the paper. 
The skill pool is distilled from training-phase trajectories(Appendix~\ref{app:skill-construction}).
\textit{SkillsBench}~\citep{li2026skillsbenchbenchmarkingagentskills} releases 87 agent tasks across 11 domains with human-curated skills.
As in prior work that uses SkillsBench, we restrict evaluation to the 69 self-contained subset whose tasks can build and run offline on our cluster. The remaining 18 tasks are listed by category in Appendix~\ref{app:skillsbench-filter}.
The 11 domains are uneven in size and individually small, so we report a single aggregate score on SkillsBench.
\textit{ALFWorld}~\citep{shridhar2021alfworld} is a text-based decision-making benchmark across six household categories.
We host the environment via AgentGym~\citep{xi-etal-2025-agentgym}. The skill pool is distilled like tau2-bench(Appendix~\ref{app:skill-construction}).

\paragraph{Implementation.}
The agent model is Qwen3-235B-A22B-Instruct-2507 in tau2-bench and SkillsBench; ALFWorld uses Qwen3-8B, held uniform across compared methods~\citep{yang2025qwen3technicalreport}.
Our planner $f_\theta$ is a simple MLP over a frozen Qwen3-Embedding-0.6B encoder~\citep{zhang2025qwen3embeddingadvancingtext}, and the renderer $R_\phi$ is Qwen3-8B fine-tuned from a Qwen3-235B-A22B-Instruct-2507 teacher under the curriculum.
Each (method, task) configuration is run five times with distinct seeds, on NVIDIA H200 GPU nodes.
Full hyperparameters are in Appendix~\ref{app:impl}.

\paragraph{Baselines.}
We compare \skillinjector{} against baselines drawn from four families, all sharing the same frozen agent and skill pool.
Boundary conditions include a no-skill lower bound, random single-skill injection that measures the harmful-injection effect, and full-library injection where pool size permits.
Classical retrieval covers BM25 as a sparse retriever and a Dense Cosine retriever that ranks skills by cosine similarity between task and skill embeddings~\citep{10.1561/1500000019}.
SkillRouter~\citep{zheng2026skillrouterskillroutingllm} reranks the candidate listwise over the skill body, and Graph of Skills~\citep{liu2026graphskillsdependencyawarestructural} retrieves via traversal over a skill-similarity graph.
The generative selector is an LLM-as-selector that lets the frozen agent backbone emit skill identifiers.
Our \skillinjector{} combines the planner with the renderer resolving selection, budget, and presentation jointly.
Per-baseline implementation details are in Appendix~\ref{app:baseline}.

\paragraph{Evaluation.}
The primary metric is task pass rate.
We use agent-level binary success on tau2-bench, the \texttt{pytest} verifier result on SkillsBench, and per-category episode success on ALFWorld.
For ablations where the harness records interaction traces, we additionally report the average number of agent messages per task, denoted $\bar{M}$, where lower values indicate lower cost at comparable task completion.
We report the mean across the five seeds in all main-text tables.
During test evaluation, methods do not update thresholds, receive execution feedback, perform online search, or re-rank skills after observing test outcomes.

\begin{figure*}[t]
\centering
\includegraphics[width=\textwidth]{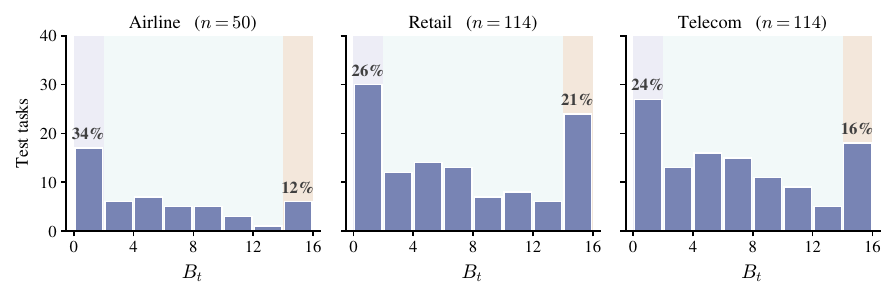}
\caption{\textbf{Adaptive budget distribution.}
Each panel shows the histogram of per-task selected budget $B_t$ in one tau2-bench domain, under the dev-calibrated $\tau^\star_d$ and cap $B_{\max} = 16$.}
\label{fig:bt-histogram}
\end{figure*}

\subsection{Main Results}
\label{sec:results:main}

Table~\ref{tab:main} reports task pass rate of \skillinjector{} against all baselines on the three benchmarks.
\skillinjector{} attains the best score in every column, with an average pass rate of 58.7\%, 5.1 points above the strongest baseline.
The consistent gap across benchmarks supports the claim that jointly optimizing selection, budget, and rendering outperforms optimizing selection alone, which is what every compared baseline does.

\subsection{Component ablation}
\label{sec:results:component}

We isolate the contribution of each part of the \skillinjector{} pipeline.
\textit{w/o renderer} reverts descriptions to their original form before injection; \textit{w/o adaptive budget} replaces the per-task budget with the best fixed top-$K$; \textit{w/o planner} injects the top-$K$ skills judged globally most useful on dev rollouts.
The no-skill condition serves as the lower bound.
Table~\ref{tab:ablation:component} reports pass rate and average agent messages per task ($\bar{M}$).

\begin{table}[h]
\centering
\scriptsize
\setlength{\tabcolsep}{2.5pt}
\begin{tabular}{lcccccc}
\toprule
& \multicolumn{2}{c}{airline} & \multicolumn{2}{c}{retail} & \multicolumn{2}{c}{telecom} \\
\cmidrule(lr){2-3}\cmidrule(lr){4-5}\cmidrule(lr){6-7}
Variant & pass\,$\uparrow$ & $\bar{M}\,\downarrow$ & pass\,$\uparrow$ & $\bar{M}\,\downarrow$ & pass\,$\uparrow$ & $\bar{M}\,\downarrow$ \\
\midrule
\textbf{Full \skillinjector{}}  & \textbf{60.0} & 28.5 & \textbf{61.4} & 25.7 & \textbf{67.0} & 61.5 \\
\quad w/o renderer              & \underline{55.2} & 36.7 & \underline{59.6} & 33.2 & \underline{65.8} & 67.6 \\
\quad w/o planner               & 47.2 & \underline{26.0} & 51.4 & \textbf{24.5} & 52.5 & 57.1 \\
\quad w/o adaptive budget       & 51.6 & 29.7 & 56.5 & 25.4 & 56.1 & \underline{56.0} \\
\midrule
No-skill                         & 37.6 & \textbf{25.4} & 51.2 & \underline{24.7} & 40.0 & \textbf{55.4} \\
\bottomrule
\end{tabular}
\caption{\textbf{Component ablation.}
The renderer prevents much of the extra interaction cost.
Arrows indicate the preferred direction.}
\label{tab:ablation:component}
\end{table}

\textbf{The planner is the dominant lever.} Removing the planner produces the largest pass-rate drop on every domain.
\textbf{The renderer pays back on both metrics.} Removing the renderer inflates $\bar{M}$ by 6.1 to 8.2 messages on every domain. Every other ablation either keeps $\bar{M}$ roughly flat or compresses it alongside pass rate.
\textbf{Much of the overhead is recovered by the renderer.} Full \skillinjector{} raises $\bar{M}$ above no-skill on every domain.
Crucially, the \textit{w/o renderer} row pushes $\bar{M}$ a further 6 to 8 messages higher while losing pass rate.

\begin{figure}[h]
\centering
\includegraphics[width=\columnwidth]{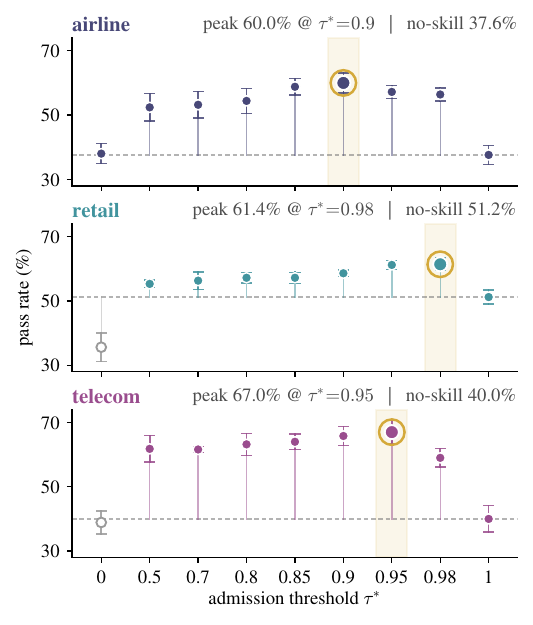}
\caption{\textbf{Admission-threshold sweep.}
Whiskers show $\pm 1\sigma$ over five seeds, the dashed line is the no-skill baseline, hollow markers denote means at or below the baseline, and the highlighted marker gives the dev-selected $\tau^\star_d$.}
\label{fig:tau-sweep}
\end{figure}

\subsection{Adaptive budget analysis}
\label{sec:results:budget}

\begin{figure*}[t]
\centering
\includegraphics[width=\textwidth]{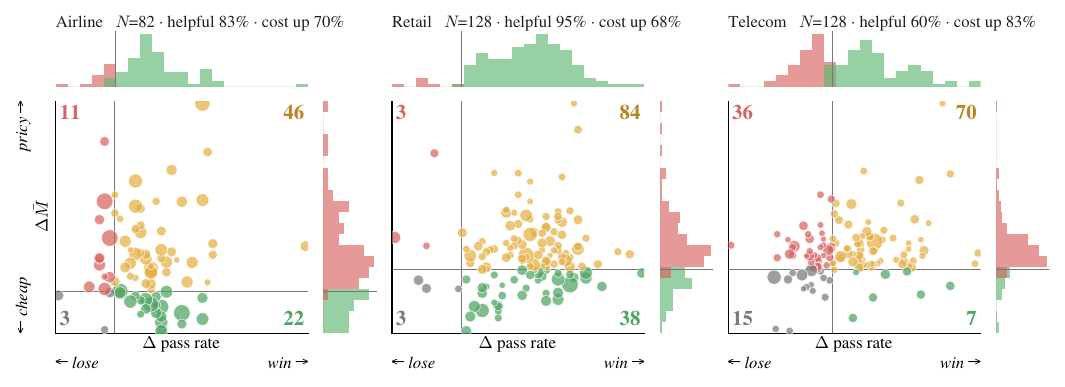}
\caption{\textbf{Per-skill effects.}
Each point is one skill, plotted by its change in pass rate and in agent messages per task against the no-skill baseline.
One panel per tau2-bench domain.
Most skills are helpful but raise interaction cost, while a long tail actively reduces pass rate.
This motivates selecting skills by predicted utility rather than injecting a domain-level shortlist.}
\label{fig:per-skill-delta}
\end{figure*}

We sweep nine admission thresholds from permissive to strict(Figure~\ref{fig:tau-sweep}).
\textbf{Both extremes collapse to the no-skill baseline.}
Accept-all ($\tau{=}0$) inherits the full-library degradation, and accept-none ($\tau{=}1$) returns to no-skill.
The gain from skill injection comes entirely from operating in between.
\textbf{The plateau between the extremes is wide.}
Pass rate stays close to its peak across a broad range of intermediate thresholds, so $\tau^\star_d$ does not need fine-grained tuning to recover most of the gain.

We then turn to the distribution of per-task budgets $B_t$ chosen by the adaptive rule (Figure~\ref{fig:bt-histogram}).
\textbf{The selected budget genuinely varies across tasks.}
The histogram shows a three-part distribution on every domain:
a left mode that rejects nearly all skills, a broad middle band that selects a moderate number, and an at-cap spike for a minority of tasks.
\textbf{The cap binds occasionally rather than inertly.}
A small fraction of tasks lands at $B_{\max}$, so the cap is a real upper bound.
The running example $t_{47}$ (Section~\ref{sec:problem}) lands at $B_{t_{47}} = 4$ in the middle band.

\subsection{Planner ablation}
\label{sec:results:planner}

Within the planner we ablate the two loss components and the encoder backbone.
\textit{w/o $\mathcal{L}_{\mathrm{align}}$} keeps only the pairwise preference signal; \textit{w/o $\mathcal{L}_{\mathrm{pref}}$} keeps only the soft alignment to the benefit distribution.
We additionally swap the 0.6B encoder for Qwen3-Embedding-4B and 8B variants to test whether further gains come from the supervision or from representation capacity.
Table~\ref{tab:ablation:planner} reports pass rate and $\bar{M}$ for each variant.

\textbf{Both loss terms matter.} 
Removing either term lowers pass rate on every domain.
Alignment shapes the global ranking across the candidate set while preference sharpens local pairwise contrasts among close skills.
\textbf{Encoder scaling yields no significant gain.} 
Replacing the 0.6B encoder with a 4B or 8B variant only slightly affects pass rate on every domain.
The planner is therefore more supervision-bottlenecked rather than representation-bottlenecked.
\textbf{Interaction cost is insensitive to planner choices.} 
$\bar{M}$ stays within roughly $\pm 2$ messages of Full across the four ablations on airline and retail.
The planner controls which skills are chosen, not how the agent then exercises them.

\begin{table}[h]
\centering
\scriptsize
\setlength{\tabcolsep}{2.5pt}
\begin{tabular}{lcccccc}
\toprule
& \multicolumn{2}{c}{airline} & \multicolumn{2}{c}{retail} & \multicolumn{2}{c}{telecom} \\
\cmidrule(lr){2-3}\cmidrule(lr){4-5}\cmidrule(lr){6-7}
Variant & pass\,$\uparrow$ & $\bar{M}\,\downarrow$ & pass\,$\uparrow$ & $\bar{M}\,\downarrow$ & pass\,$\uparrow$ & $\bar{M}\,\downarrow$ \\
\midrule
\textbf{Full}                                  & \textbf{60.0} & 28.5 & \textbf{61.4} & 25.7 & \textbf{67.0} & 61.5 \\
\quad w/o $\mathcal{L}_{\mathrm{align}}$       & 53.2 & 28.6 & 57.7 & \textbf{24.8} & 62.3 & \textbf{56.7} \\
\quad w/o $\mathcal{L}_{\mathrm{pref}}$        & 54.4 & 29.3 & 56.1 & 25.2 & 64.6 & \underline{58.7} \\
\quad encoder $\to$ 4B                         & 58.0 & \textbf{27.0} & 58.9 & 26.7 & \underline{65.1} & 61.3 \\
\quad encoder $\to$ 8B                         & \underline{58.8} & \underline{27.5} & \underline{59.5} & \underline{24.9} & 64.0 & 67.6 \\
\bottomrule
\end{tabular}
\caption{\textbf{Planner ablation.}
Both alignment and preference losses are needed, and larger encoders yield no meaningful gain.}
\label{tab:ablation:planner}
\end{table}

\begin{figure*}[t]
\centering
\includegraphics[width=\textwidth]{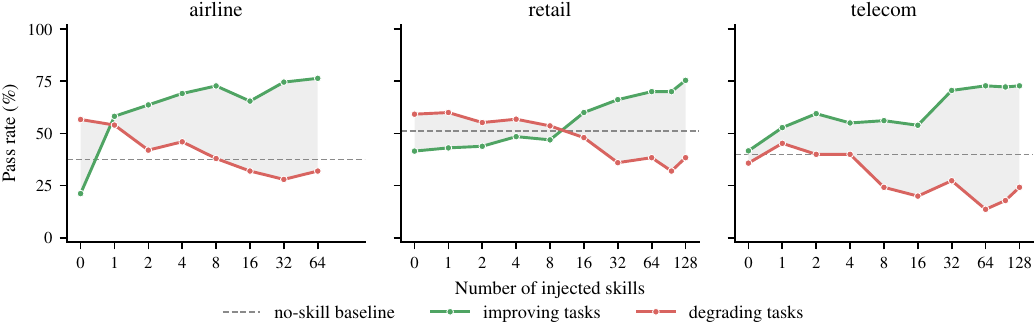}
\caption{\textbf{Per-task responses to larger skill sets.}
Each panel shows the average pass-rate trajectory of two task groups within one tau2-bench domain.
Tasks are grouped by the Spearman correlation between $N$ and reward into improving ($\rho > 0.3$) and degrading ($\rho < -0.3$).
The two groups diverge as $N$ grows, showing why a fixed large skill budget fails at the aggregate level.}
\label{fig:per-task-curve}
\end{figure*}

\subsection{Renderer ablation}
\label{sec:results:renderer}

We contrast our curriculum against simpler renderer training regimes and against the 235B teacher used during distillation.
\textit{8B prompt only} runs the same 8B base model without any fine-tuning; \textit{8B trace-rich only} and \textit{8B trace-free only} keep the curriculum's data sources but apply only one of the two regimes.
The 235B teacher provides a soft upper bound, and the no-render condition is the lower bound on rendering quality.

\textbf{The small renderer requires training, not just prompting.}
Prompt-only 8B drops below the no-render baseline on all three domains, showing that set-aware adaptation must be learned rather than prompted only.
\textbf{Single-regime distillation helps but is insufficient.} 
The trace-rich and trace-free solo variants both recover the gap. The curriculum schedule itself contributes to the final renderer.
For example, for $t_{47}$ the trace-rich variant of \texttt{cancellation\_eligibility\_criteria} adds trace-conditional gates like \texttt{Do NOT call if the user has already been informed of the policy}, while the trace-free variant produces generic guards like \texttt{Do NOT call if the reservation ID is missing}.
The curriculum-trained renderer instead names sibling skills directly, for instance \texttt{Do NOT call if the user is asking about a change, refund}.
\textbf{The curriculum-trained 8B tracks the 235B teacher at a fraction of the cost.}
The 8B curriculum approaches the teacher's pass rate, and it also reduces a little agent interaction cost beyond the inference savings.

\begin{table}[h]
\centering
\scriptsize
\setlength{\tabcolsep}{2.5pt}
\begin{tabular}{lcccccc}
\toprule
& \multicolumn{2}{c}{airline} & \multicolumn{2}{c}{retail} & \multicolumn{2}{c}{telecom} \\
\cmidrule(lr){2-3}\cmidrule(lr){4-5}\cmidrule(lr){6-7}
Renderer variant & pass\,$\uparrow$ & $\bar{M}\,\downarrow$ & pass\,$\uparrow$ & $\bar{M}\,\downarrow$ & pass\,$\uparrow$ & $\bar{M}\,\downarrow$ \\
\midrule
No render                  & 55.2 & 36.7 & 59.6 & 33.2 & 65.8 & 67.6 \\
8B prompt only             & 49.6 & \textbf{28.5} & 52.1 & \textbf{25.6} & 47.4 & 62.1 \\
8B trace-rich only         & 53.6 & 33.9 & 57.4 & 26.1 & 64.7 & 84.4 \\
8B trace-free only         & 52.4 & 36.5 & 56.8 & 27.3 & 63.2 & \textbf{58.7} \\
235B teacher               & \textbf{62.4} & 29.3 & \underline{60.5} & 39.9 & \textbf{70.7} & \underline{61.3} \\
8B curriculum(Ours)     & \underline{60.0} & \textbf{28.5} & \textbf{61.4} & \underline{25.7} & \underline{67.0} & 61.5 \\
\bottomrule
\end{tabular}
\caption{\textbf{Renderer ablation.}
The curriculum-trained 8B renderer approaches teacher-level pass rate.}
\label{tab:ablation:renderer}
\end{table}

\subsection{Additional Results}
\label{sec:results:additional}

We further analyze two diagnostic views behind the macro results.
Figure~\ref{fig:per-skill-delta} resolves the macro pass rate into per-skill effects on both quality and cost.
\textbf{Skills are heterogeneous in utility.}
It shows that a subset of skills actively lowers pass rate even though they look topically relevant, and many useful skills also raise the message budget.
The planner must therefore select on predicted utility along both axes, not on surface similarity.
Figure~\ref{fig:per-task-curve} traces per-task pass-rate trajectories as the injected skill budget grows.
\textbf{Tasks are heterogeneous in skill demand.}
As more skills are injected (Figure~\ref{fig:per-task-curve}), task groups with different sensitivities to skills diverge, with the gap widening to tens of percentage points.
This is what task-adaptive budgeting must serve.
See Appendix~\ref{app:case-study} for more details about case study.

\section{Conclusion}
\label{sec:conclusion}

This paper shows that skill injection is not only a retrieval problem, but a question of constructing the skill context a frozen agent can read.
\skillinjector{} implements this view in two ways. First, it uses utility signal to plan the exposed skill set. Second, it renders selected descriptions in a set-aware form.
Across tau2-bench, SkillsBench, and ALFWorld, it improves task completion over many retrieval baselines. The ablations show that selection, budget, and presentation each affect downstream performance.
These results indicate that a skill's value is conditional on the task and other skills sharing the context.
The primary object to optimise is the context the agent actually consumes.
A natural next step is to co-train the skill library and the injection layer under a shared task-utility objective.

\section*{Limitations}

\skillinjector{} is evaluated as an inference-time layer for frozen agents on tau2-bench, SkillsBench, and ALFWorld.
These benchmarks already span dialogue-driven tool use, packaged coding skills, and text-based interactive decision-making, but additional task families such as GUI automation, open-web agents, and very long-horizon embodied tasks remain untested~\citep{NEURIPS2024_5d413e48, ICLR2024_4410c071}.
Within each benchmark we use a single source of skills, either human-curated or LLM-generated, and do not study mixed pools.
Cost considerations restrict our agent backbones to the open Qwen series, so our results do not characterize performance under proprietary closed-source models.
The frozen-agent assumption matches most current deployments where the policy is held fixed at inference, and we leave the interaction with co-trained or RL-finetuned agents to follow-up work.

\section*{Ethical Considerations}

\skillinjector{} runs at inference time on a frozen LLM agent. We do not release a new dataset or change the agent's policy. The risks we describe are inherited from the components we build on.


\bibliography{custom}

\appendix
\section{Case Study}
\label{app:case-study}

This appendix collects the reference data for $t_{47}$, the airline cancellation task used as the running example throughout this paper.
All numbers and rewrites are drawn from the same experimental runs as Table~\ref{tab:main}.

\subsection{Task and library}
\label{app:case-study:task}

The task asks the agent to refuse a flight-cancellation request because the user's stated reason (a friend's birthday) is not covered by the airline's insurance policy.
It is drawn verbatim from tau2-airline (task id 47).

\begin{Verbatim}[fontsize=\footnotesize,breaklines=true]
purpose:         Check that the agent understands that insurance only covers health or weather reasons for cancellation.
reason_for_call: You want to cancel your flight because the flight coincides with your best friend's birthday.
baseline pass:   0.60 (3/5 trials with no skill injected)
\end{Verbatim}

The skill library $\mathcal{S}$ contains $82$ airline policy-reference skills, detailed in Appendix~\ref{app:skill-construction}.

\subsection{Planner scores}
\label{app:case-study:planner}

For illustration, we present five representative candidates from the $82$-skill airline library and adopt an admission threshold of $\tau = 0.80$ for this case study.
Table~\ref{tab:case:planner} reports the trained planner $f_\theta$ scores and the resulting admission decisions.
Four skills are admitted to $\mathcal{C}^*_t$, and one is rejected.

\begin{table}[h]
\centering
\scriptsize
\setlength{\tabcolsep}{4pt}
\begin{tabular}{lcc}
\toprule
Skill & $f_\theta$ score & Decision \\
\midrule
\texttt{cancellation\_eligibility\_criteria} & $0.94$ & admit \\
\texttt{cancellation\_eligibility\_policy}   & $0.92$ & admit \\
\texttt{refund\_policy\_reference}           & $0.86$ & admit \\
\texttt{insurance\_coverage\_reference}      & $0.81$ & admit \\
\texttt{modify\_flight\_insurance\_policy}   & $0.39$ & reject \\
\bottomrule
\end{tabular}
\caption{\textbf{Candidate skills for $t_{47}$.}
Trained planner scores and admission decisions. \texttt{modify\_flight\_insurance\_policy} shares the ``insurance'' keyword with the user's wording, but its body returns post-booking insurance guidance rather than cancellation coverage.}
\label{tab:case:planner}
\end{table}

\subsection{Adapted descriptions}
\label{app:case-study:rewrites}

For each admitted skill, Table~\ref{tab:case:rewrites} reports the original description $d_i$ and the scope clause that the trained renderer appends to produce $\tilde{d}_i$.
Each appended clause names at least one of the three co-selected skills.
Although the two \texttt{cancellation\_*} skills share a near-identical name, the renderer-added scopes carve out non-overlapping functions, with \texttt{\_criteria} covering cancellation eligibility itself and \texttt{\_policy} restricted to refund-eligible cancellation.

\subsection{Renderer variants}
\label{app:case-study:variants}

For \texttt{cancellation\_eligibility\_criteria}, the four renderer variants of Table~\ref{tab:ablation:renderer} produce different prompt-guidance strings on this task.
The full strings are reproduced below verbatim.

\section{Method Details}
\label{app:method-details}

This appendix expands on the design choices behind the two trained components in Section~\ref{sec:method}.
Algorithm~\ref{alg:pipeline} gives the full offline training and test-time injection pipeline.
We then discuss why the planner uses a two-term loss and how its candidate set yields informative negatives.
Then we will describe how the set-aware renderer encodes neighbor context through supervision.

\subsection{Planner}
\label{app:methdet:planner}

\paragraph{Why two loss terms.}
The alignment term $\mathcal{L}_{\mathrm{align}}$ pulls $p_\theta(\cdot \mid t)$ toward the full benefit distribution $q_\Delta(\cdot \mid t)$.
Because KL is averaged over the simplex, two skills that the data ranks in opposite directions but with nearby benefit scores receive little corrective signal from this term alone.
The preference term $\mathcal{L}_{\mathrm{pref}}$ acts on per-pair odds ratios, so the local contrasts that $\mathcal{L}_{\mathrm{align}}$ averages out remain in the gradient.
We adopt the odds-ratio because it does not require a reference policy and composes additively with $\mathcal{L}_{\mathrm{align}}$ in the same probability space~\citep{hong-etal-2024-orpo, molfetta-etal-2025-ports}.
With $\lambda = 0.3$, the two terms cooperate rather than compete. 
$\mathcal{L}{\mathrm{align}}$ governs the distributional shape, whereas $\mathcal{L}{\mathrm{pref}}$ encourages separation among the top contenders.

\paragraph{Candidate set and hard negatives.}
The formalism in Section~\ref{sec:method:planner} writes $\mathcal{S}_c(t) \subseteq \mathcal{S}$ to leave room for a coarse pre-filter when the pool is large.
In the libraries used here $\lvert \mathcal{S} \rvert$ is at most a few hundred skills per domain (Appendix~\ref{app:skill-construction}), and scoring a single $(t, s)$ pair through the frozen Qwen3-Embedding-0.6B encoder and a small MLP head is inexpensive~\citep{zhang2025qwen3embeddingadvancingtext}.
So we set $\mathcal{S}_c(t) = \mathcal{S}$ throughout this paper and let the planner rank the full per-domain pool for every task.
This choice avoids a recall ceiling imposed by an upstream retriever and keeps the hard-negative pool global.
Within $\mathcal{S}$, skills that are semantically close to a preferred candidate but receive a lower $\Delta$ are treated as hard negatives for $\mathcal{L}_{\mathrm{pref}}$. 
This encourages the gradient to separate functionally similar skills, rather than spending capacity on topically unrelated pairs.
For label sharpness and predicted softmax we use $\beta = 0.5$ and $\gamma = 0.5$, both chosen by validation utility (Appendix~\ref{app:impl}).

\subsection{Renderer}
\label{app:methdet:renderer}

\paragraph{Where the set context lives.}
The renderer reads, at both training and inference, the target description $d_i$, the task $t$, and the set of co-selected neighbor descriptions $N = \{d_j : s_j \in \mathcal{C}^*_t \setminus \{s_i\}\}$.
Therefore the rewrite of $d_i$ can disambiguate against the specific neighbors it will be co-injected with.
During training the teacher additionally consumes a trace summary $h_i$ collected from a training rollout of $s_i$, which exposes behavioral evidence about why $d_i$ should be changed.
This trace block is also present in the student's trace-rich input but is dropped from the trace-free input.
The student therefore inherits trace-grounded supervision while learning to produce the same set-aware target from descriptions alone at inference.
Table~\ref{tab:render-conditioning} summarises the design axes.

\paragraph{Curriculum schedule.}
The student is trained over two consecutive epochs that share optimizer state and a single cosine schedule.
At each step we mix trace-rich and trace-free inputs by a per-sample Bernoulli draw. 
Each skill's record is independently replaced by its trace-free counterpart with probability $\rho_k$, where $\rho_1 = 0.1$ in the first epoch and $\rho_2 = 0.9$ in the second.
Stage one therefore exposes $R_\phi$ mostly to trajectory-grounded rewrites, anchoring the rewrite policy on behavioral evidence, while stage two shifts the distribution toward the trace-free input that the renderer will face at inference.
Because the mix is a Bernoulli draw rather than a hard switch, no epoch contains only one regime, and the cross-stage transition does not require a learning-rate reset.
At inference $R_\phi$ reads only the skill description fields with the trace block omitted. The set-awareness it produces in $\tilde{d}_i$ is the structure it learned to predict from $D_2$.

\section{Implementation Details}
\label{app:impl}

\subsection{Planner and Renderer Training}

We describe the training settings of the two components on tau2-bench, then note how ALFWorld and SkillsBench reuse the same pipeline.

\paragraph{Planner $f_\theta$.}
The encoder Qwen3-Embedding-0.6B is kept frozen. Only a three-layer MLP head with hidden width $128$ and dropout $0.1$ consumes the concatenated $(t, s)$ embedding and outputs a logit.
We optimize the head with Adam at learning rate $10^{-3}$, no weight decay, and no warmup. Updates are taken on single-step batches that pack all in-batch pairs through a dedicated pair builder.
Mixed precision is disabled and the head trains in fp32.
The dual loss uses $\beta = 0.5$ on the $\Delta$ labels, $\gamma = 0.5$ on the predicted softmax, and weight $\lambda = 0.3$ on the preference term (Eq.~\ref{eq:align} for $\mathcal{L}_\mathrm{align}$; $\mathcal{L}_\mathrm{pref}$ is given inline in Section~\ref{sec:method:planner}).
After training, the per-domain threshold $\tau^\star_d$ is calibrated on a held-out dev set under per-task min--max normalization, yielding $\tau^\star = 0.90 / 0.98 / 0.95$ for airline / retail / telecom.
At evaluation each task is run with five seeds ($300, 301, 302, 303, 304$).

\paragraph{Renderer $R_\phi$.}
SFT runs through Hugging Face Trainer with AdamW at learning rate $5{\times}10^{-5}$, cosine schedule with warmup ratio $0.10$, per-device batch size $1$, gradient accumulation $8$ (effective batch $8$), bf16, gradient checkpointing, and the SDPA attention kernel.
The maximum combined input--output length is $8{,}192$ tokens.
Training spans two epochs that share optimizer state. The trace-rich / trace-free mix uses $\rho_1 = 0.1$ in epoch one and $\rho_2 = 0.9$ in epoch two (Appendix~\ref{app:methdet:renderer}).
The teacher Qwen3-235B-A22B-Instruct-2507 used to produce $D_2$ is served through the same SGLang stack at supervision time.

\paragraph{Cross-benchmark settings.}
ALFWorld and SkillsBench reuse the same planner and renderer pipelines as tau2-bench.
The only nominal differences are the choice of training task pool (Appendix~\ref{app:skill-construction}) and per-domain $\tau^\star$ recalibrated on each benchmark's held-out dev set.
Decoding and curriculum settings stay at the values reported above.

\paragraph{Compute.}
We train and evaluate on NVIDIA H200 GPUs.
The aggregate compute reported below is dominated by benchmark inference, which is shared across all compared methods rather than attributable to \skillinjector{}. Across the three benchmarks, all $K$ compared methods are run with $5$ seeds each, totaling approximately $240$ H200 GPU-hours. Qwen3-235B-A22B-Instruct-2507 (tau2-bench, SkillsBench) and Qwen3-8B (ALFWorld) are served as the frozen agent backbones for every method, and this cost would arise under any retrieval or skill-routing baseline evaluated on the same protocol. Training the planner head (a 3-layer MLP over a frozen Qwen3-Embedding-0.6B encoder) and distilling the 8B renderer from the Qwen3-235B-A22B-Instruct-2507 teacher together add approximately $24$ H200 GPU-hours, dominated by the one-off teacher generation of $D_2$. At inference, \skillinjector{} adds only the MLP planner and a single 8B renderer call on one H200 on top of the shared agent backbone; the 235B model is used only once, offline, to produce the renderer's supervision targets.

\subsection{Per-benchmark Evaluation Protocol}

\paragraph{tau2-bench.}
We use Qwen3-235B-A22B-Instruct-2507 as the frozen agent backbone, served through SGLang.
Generation runs with temperature 0.3, \texttt{max\_tokens}=4096, and no-thinking mode. 
Each rollout is capped at 50 environment steps; trajectories that exceed this limit are recorded as max-step terminations. 
We run five trials per (task, method) pair.

\paragraph{SkillsBench.}
The agent backbone, generation parameters, step limit, trial count, and seed protocol are identical to tau2-bench.
Each trial is executed inside an isolated Docker container, and the upstream \texttt{pytest} verifier is invoked at the end of the trial to produce a binary reward.

\paragraph{ALFWorld.}
The environment is hosted on AgentGym through the AgentGym EnvServer with HTTP endpoints \texttt{/create}, \texttt{/reset}, and \texttt{/step}.
The evaluation split covers item ids 2420--2619 (200 episodes total), distributed across six task families: \texttt{pick\_and\_place\_simple} (46), \texttt{pick\_two\_obj\_and\_place} (45), \texttt{pick\_clean\_then\_place\_in\_recep} (37), \texttt{pick\_cool\_then\_place\_in\_recep} (28), \texttt{pick\_heat\_then\_place\_in\_recep} (25), and \texttt{look\_at\_obj\_in\_light} (19).
The agent backbone is Qwen3-8B, supervised-fine-tuned on AgentGym's public AgentTraj-L Round-0 expert traces (14{,}485 examples; learning rate $10^{-5}$, one epoch) to ensure basic environment competence. 
We serve it on a single H200 GPU through SGLang 0.5.9 with context length 32{,}768, \texttt{mem\_fraction\_static}=0.85, and tensor-parallel size 1, again in no-thinking mode. 
Each task is allowed up to 30 interaction rounds (\texttt{MAX\_ROUND}=30). We run five trials per task, all sampled at temperature 0.7 and seeded from a fixed base seed of 42.

\section{Prompts}
\label{app:prompts}

This appendix lists the three prompts that materially shape the \skillinjector{} pipeline: the teacher prompt used by the strong distiller to produce set-aware rewrite targets $D_2$ for the renderer (Prompt~1), the SFT input template used by the student renderer at training time (Prompt~2), and the prompt used by the LLM-as-selector baseline (Prompt~3).

\noindent The LLM-as-selector baseline is served by Qwen3-235B-A22B-Instruct-2507 with temperature $0$, $\texttt{max\_tokens}=64$, and a single user message; \texttt{skill\_list} is the newline-joined list of \texttt{- <name>: <short description>} entries truncated to 300 characters per skill.

\section{Baselines}
\label{app:baseline}

All baselines share the same frozen agent, candidate skill pool, and evaluation harness on every benchmark. 
Only the skill-selection rule changes between rows of Table~\ref{tab:main}.

\paragraph{Boundary conditions.}
The \textit{no-skill} baseline injects no skill context and serves as the lower bound. The \textit{random-skill} baseline samples one skill uniformly at random from the candidate pool per task, measuring whether arbitrary skill context is preferable to none. The \textit{full-library} baseline injects every skill in the candidate pool and measures the cost of saturation.

\paragraph{Similarity-based retrievers.}
BM25 is implemented with the \texttt{rank\_bm25} library using each task instruction as query and the textual skill content (one document per skill) as corpus~\citep{10.1561/1500000019}. Dense Cosine encodes both the task instruction and each skill text with a frozen Qwen3-Embedding-0.6B encoder~\citep{zhang2025qwen3embeddingadvancingtext}, applies $L_2$ normalisation, and ranks skills by inner product. Both retrievers score every candidate independently and share the same top-$K$ budget as the other retrieval baselines.

\paragraph{Structure-aware retrievers.}
SkillRouter~\citep{zheng2026skillrouterskillroutingllm} is a listwise reranker that scores the full candidate set jointly conditioned on the task. Graph of Skills~\citep{liu2026graphskillsdependencyawarestructural} pre-builds a similarity graph over the skills and retrieves through graph traversal seeded by the task representation. We re-implement both methods locally; training data, scorer architecture, and remaining hyperparameters follow the originals' published recipes.

\paragraph{Generative selector.}
The LLM-as-selector baseline prompts the frozen agent backbone with the task instruction and a structured list of available skills (identifier plus short description) and asks it to emit the identifier of the skill it would invoke. We use few-shot prompting and apply no fine-tuning to the backbone. The prompt template is in Appendix~\ref{app:prompts}.

\section{Skill Library Construction}
\label{app:skill-construction}

\paragraph{tau2-bench.}
The skill library for each tau2-bench domain is distilled from baseline rollouts of the un-augmented agent in four steps.
First, we collect three-trial baseline rollouts (seeds $300, 301, 302$).
For each task we format the three trajectories into a single transcript with PASS / FAIL markers and label the task ALL\_PASS, ALL\_FAIL, or MIXED based on the reward vector.
Second, we prompt the teacher Qwen3-235B-A22B-Instruct-2507 once per task with the domain policy, the tool list, the task description and evaluation criteria, the formatted trajectories, one positive seed example (\texttt{cancellation\_guard}), one negative seed example (\texttt{action\_executor}), and a type-specific analysis instruction. 
The teacher emits 1--3 candidate skill modules per call, each carrying a name, description, prompt guidance, target pattern, and a reference-style function body.
Each task is sampled at five temperatures $\{0.0, 0.3, 0.5, 0.7, 1.0\}$, yielding $250 / 570 / 570$ raw generations for airline / retail / telecom respectively.
Third, we embed every generated skill with Qwen3-Embedding-8B over the concatenated name, description, and docstring, cluster by cosine threshold $0.90$, and rank within each cluster by within-cluster frequency, task-coverage breadth, and MMR diversity, leaving $86 / 154 / 140$ candidates per domain.
Fourth, light human curation rejects only modules that fail to load, use action-semantic names (\texttt{executor}, \texttt{router}, \texttt{handler}, \texttt{processor}), or copy \texttt{policy.md} verbatim.
Semantically near-duplicate skills are intentionally retained so that collision density emerges naturally at evaluation time.
The libraries used in the scaling experiments are then capped at $82 / 128 / 128$ skills per domain by MMR rank, matching the largest $K$ explored in Figure~\ref{fig:scaling}.

\paragraph{SkillsBench.}
SkillsBench ships with a curated skill folder per task, originally intended for an oracle-injection evaluation that grants each task privileged access to its own paired skills.
We convert this to an open-pool retrieval setting by aggregating across tasks.
After the $87 \to 69$ task filter described in Appendix~\ref{app:skillsbench-filter}, we walk each surviving task's \texttt{environment/skills/} directory and unify entries by the sha8 hash of their canonicalized body.
Identical bodies appearing under multiple tasks are merged into a single pool entry that records all source-task IDs.
Near-identical content variants under the same display name are kept as separate entries so that the pool reflects authoring drift rather than name collisions.
The aggregated open pool contains $150$ unique skill bodies across $142$ display names ($8$ names carry more than one content variant), drawn from $176$ \texttt{SKILL.md} occurrences in total.
All retrieval-based methods, including our planner, query this global pool with only the task instruction.
No method receives privileged access to its own training-paired skills.

\paragraph{ALFWorld.}
The skill pool is distilled from the ALFWorld subset of AgentGym's AgentTraj-L corpus~\citep{xi-etal-2025-agentgym}, which contains $2{,}420$ expert trajectories spanning the six task families.
A Qwen3-235B-A22B-Instruct-2507 distiller converts these trajectories into $1{,}920$ raw candidate skills,  following the same procedure used for tau2-bench. Then we compress the candidate set with a three-stage filter.
First, we remove exact duplicates using the sha8 hash of the canonicalized skill body.
Second, we perform semantic clustering with Qwen3-Embedding-8B.
We use the threshold ladder $0.90 \to 0.85 \to 0.80$ and retain the cluster medoid at each step.
Third, we apply a budget gate based on the Qwen3-8B tokenizer and the actual injection template.
This gate enforces a hard $24{,}000$-token context budget while keeping at least two skills per task family.

The pipeline therefore mirrors the tau2-bench construction protocol in steps and tooling. It differs only in the source corpus and in the budget-gate stage, which is adapted to ALFWorld's longer in-context demands.

\section{SkillsBench Task Selection}
\label{app:skillsbench-filter}

SkillsBench~\citep{li2026skillsbenchbenchmarkingagentskills} contains 87 self-rated tasks.
We found that 18 of these tasks could not be reproduced offline within our containerised infrastructure.
These tasks were excluded from the evaluation.
Table~\ref{tab:skillsbench-filter} reports the 18 excluded tasks, grouped by failure mode.
The first two categories are Docker build failures and pre-build failures.
In both cases, the failures are caused by dependencies in the upstream Dockerfiles.
Some build steps require external mirrors, registries, or version-specific toolchains.
These resources were not reliably accessible in our environment.
The remaining category consists of non-self-contained tasks.
These tasks require live external services at execution time.
Such services include public APIs, paid third-party endpoints, online model registries, and remote GPU resources.
The exclusion criteria were fixed before evaluation.
They were applied uniformly to all compared methods.
Thus, all methods were evaluated on the same 69-task subset.
This preserves the fairness of the benchmark comparison.

\end{document}